\documentclass[11pt]{article}

\usepackage[preprint]{acl}
\usepackage{todonotes}
\usepackage{times}
\usepackage{latexsym}
\usepackage[title]{appendix}
\usepackage[T1]{fontenc}
\usepackage[utf8]{inputenc}

\usepackage{microtype}

\usepackage{inconsolata}

\usepackage{graphicx} 
\usepackage{float} 

\usepackage{amsmath}
\usepackage{amsfonts}
\usepackage{amssymb}
\usepackage{booktabs}
\usepackage{multirow}
\usepackage{array}
\usepackage{xcolor}

\usepackage{subcaption}
\usepackage{listings}
\title{D-Mem: A Dual-Process Memory System for LLM Agents}

\author{Zhixing You\thanks{Corresponding author} \\
  Einstein Institute of Mathematics \\
  The Hebrew University of Jerusalem, Israel \\
  \texttt{zhixing.you@mail.huji.ac.il} \\\And
  Jiachen Yuan \\
  Independent Researcher \\
  \texttt{yuan.jachen@gmail.com} \\\AND
  Jason Cai \\
  AWS AI \\
  \texttt{cjinglun@amazon.com} \\}
\begin{document}
\maketitle

\begin{abstract}
  Driven by the development of persistent, self-adapting autonomous agents, 
  equipping these systems with high-fidelity memory access for long-horizon reasoning has emerged as a critical requirement.
  However, prevalent  retrieval-based memory frameworks 
  often follow an incremental processing paradigm that 
  continuously extracts and
updates conversational memories into vector databases,
relying on semantic retrieval when queried.
  While this approach is fast,
  it inherently relies on lossy abstraction,
  frequently missing contextually critical information
  and struggling to resolve queries that rely on fine-grained contextual understanding.
  To address this, we introduce D-Mem, a dual-process
  memory system. It retains lightweight vector retrieval 
  for routine queries while establishing an exhaustive Full 
  Deliberation module as a high-fidelity fallback.
  To achieve cognitive economy without sacrificing
  accuracy, D-Mem employs a Multi-dimensional Quality Gating policy to
  dynamically bridge these two processes.
  Experiments on the LoCoMo and RealTalk benchmarks using
  GPT-4o-mini and Qwen3-235B-Instruct demonstrate the
  efficacy of our approach. Notably, our Multi-dimensional Quality Gating
  policy achieves an F1 score of 53.5 on LoCoMo with GPT-4o-mini.
  This outperforms our static
  retrieval baseline, Mem0$^\ast$ (51.2), and recovers 96.7\% of the
  Full Deliberation's performance (55.3), while incurring
  significantly lower computational costs.  \footnote{Code will be released upon acceptance of the paper.}
\end{abstract}

\section{Introduction}
The development of LLM-based autonomous agents marks a significant evolution, shifting the application focus from stateless text generators to persistent, self-adapting entities deployed in dynamic, long-horizon environments
\cite{xi2023rise, wang2024survey, wang2026emem}. As agents operate
in increasingly complex scenarios, the capability to
accumulate experiences and self-evolve across extended interactions
becomes a critical necessity. However, this continuous adaptation is fundamentally
hindered by the static nature of deployed LLM parameters.
Despite the advent of ever-expanding context windows,
simply feeding massive dialogue histories into the model is computationally expensive and frequently exacerbates the "Lost-in-the-Middle" phenomenon
\cite{liu2024lost}. To address this,
recent frameworks like MemoryBank, Zep and Mem0
developed an incremental memory processing paradigm \cite{zhong2023memorybank, zep2024, mem02025}. 
These systems
continuously extract, compress, and update
conversational memories into vector databases to
enable efficient, scalable semantic retrieval.

While highly effective for routine, explicit
queries, this paradigm introduces a critical
vulnerability for deep reasoning: lossy abstraction \cite{zhu2026}.
By aggressively performing query-agnostic 
compression—condensing continuous dialogues into 
semantic snippets independent of future 
queries—the system strips away potentially crucial 
contextual nuances. Consequently, when faced 
with queries
requiring rigorous deduction, it frequently misses
contextually critical information—such as unstated
temporal logic (e.g., relative time calculations) or
multi-hop dependencies. This leaves agents struggling with
a severe performance bottleneck: static retrieval
simply cannot reconstruct the precise logical
chains lost during compression. To address the fundamental
limitations of query-agnostic memory compression, we
introduce D-Mem, a dual-process memory system that
emulates the human cognitive process of metacognitive
monitoring \cite{kahneman2011thinking, evans2013dual}.
Rather than abandoning efficient vector
retrieval, D-Mem leverages it as a rapid System 1 baseline (Mem0$^\ast$), while
introducing a robust System 2 fallback mechanism. D-Mem operates
on two distinct cognitive levels:
\begin{itemize}
  \item System 1 (Mem0$^\ast$): 
  Acting as an enhanced incremental memory processing module built upon the Mem0 architecture, this system
   performs a
        rapid, low-cost static retrieval (Top-K) based on
        surface similarity within the compressed memory space.
  \item System 2 (Full Deliberation): When
        System 1 fails to resolve the query, the system
        escalates to an exhaustive deliberate reading mode. Bypassing the
        compressed memory, this module directly processes the raw dialogue
        history. It executes a query-guided temporal scan,
        utilizing the original question as a discriminative
        anchor to systematically extract query-relevant facts
        chunk-by-chunk via a quantitative scoring mechanism (0-10). 
        By applying strict
        multi-stage filtering, it synthesizes a highly grounded, high-fidelity
        answer-closely mimicking the human cognitive
        process of purposeful reading.
\end{itemize}
Crucially, executing
System 2 indiscriminately incurs massive computational overhead. On the LoCoMo dataset,
it increases input tokens and inference time by
over 10$\times$ compared to the fast path (see Table~\ref{tab:summary_overall} for details). To achieve
cognitive economy, D-Mem incorporates a metacognitive
assessment step. Acting as a rigorous analytical 
gatekeeper, this module evaluates the initial System 1 
output against a strict multi-dimensional pass/fail 
rubric—encompassing relevance, faithfulness \& consistency,
and completeness. The architecture escalates to the 
computationally intensive System 2 exclusively when 
the initial output fails to satisfy any of these criteria, 
ensuring that Full Deliberation is 
triggered only by critical information deficits.

In summary,
our main contributions are as follows:
\begin{itemize}
  \item 
  The dual-process D-Mem Framework.
  To overcome the fundamental limitations of
  query-agnostic memory compression, we propose D-Mem,
  a novel architecture that integrates efficient
  vector retrieval (System 1) with an exhaustive
  deliberate reading mode (System 2). To
  bridge these processes and balance cognitive economy
  with reasoning accuracy, we introduce a
  Multi-dimensional Quality Gating policy (hereafter, Quality Gating). 
  Serving as a metacognitive checkpoint, it offers an
  accuracy-efficiency trade-off.

  \item Full Deliberation as a robust baseline. We 
  establish a Full Deliberation method that exhaustively 
  processes raw dialogue history chunk-by-chunk. Because 
  this extraction mechanism is strictly query-guided, 
  it ensures that explicit, nuanced details are meticulously 
  preserved. Furthermore, its chunk-by-chunk processing 
  effectively mitigates the "Lost-in-the-Middle" phenomenon. 
  By establishing this high-fidelity upper bound, 
  we isolate a critical bottleneck shift: to surpass 
  this baseline, future long memory architectures must 
  evolve beyond extracting explicit facts to actively 
  capturing and synthesizing implicit information across 
  longitudinal history.
  \item
    High Performance with Computational Efficiency. Through
    comprehensive evaluations on the LoCoMo \cite{maharana2024evaluating} and RealTalk \cite{lee2025realtalk}
    benchmarks using GPT-4o-mini and Qwen3-235B-Instruct,
    we demonstrate the efficacy of our framework. Notably,
    our Quality Gating achieves an F1 score of 53.5
    on LoCoMo and exhibits consistent substantial gains on RealTalk.
    This outperforms our Mem0$^\ast$ (51.2) and 
    recovers 96.7\% of the
    Full Deliberation's performance (55.3)
    with significantly fewer tokens and inference time.
\end{itemize}

\section{Related Work}

\subsection{Retrieval-Augmented Generation (RAG)}
Standard RAG retrieves documents via dense similarity
and feeds them to a generator. Subsequent work has
improved this pipeline through query
expansion~\citep{query_expansion},
re-ranking~\citep{reranking}, relevance
filtering~\citep{filtering}, and
query-transformation~\citep{reprompting}.
These techniques refine \textit{how} to
retrieve.
Recent \textit{Agentic RAG} paradigm
actively assess their own information needs,
refines \textit{whether} retrieval is sufficient
\citep{asai2024, activerag}.

\subsection{Agentic Memory for Autonomous Agents}
Early frameworks like MemGPT \citep{packer2023memgpt}
introduced operating system-inspired virtual context
management via hierarchical paging.
Other memory systems
like Mem0 \citep{mem02025} and MemoryBank
\citep{zhong2023memorybank} employ an incremental
processing paradigm that continuously extracts and
updates conversational memories into vector databases.
Such aggressive vectorization inherently fragments semantic context.
To preserve relational integrity, recent works have explored structural
optimizations. For instance,
G-Memory \citep{zhang2025gmemory} and Mem0${}^{g}$\citep{mem02025}
construct graph-based hierarchies,
and A-Mem \citep{amem} utilizes interconnected notes.
Despite these structural enhancements, these methods remain
fundamentally query-agnostic. Because the memory representation
is aggressively compressed and pre-fixed during the storage phase,
they inevitably suffer from \textit{lossy abstraction}—stripping
away nuanced temporal and causal logic.

To circumvent the pitfalls of lossy abstraction,
recent advancements like GAM \citep{yan2025gam}
and E-mem \citep{wang2026emem} entirely abandon query-agnostic compression.
Instead, they pivot towards episodic context reconstruction and multi-agent
deliberative paradigms, prioritizing reasoning fidelity by exhaustively
processing raw, uncompressed historical context.
However, discarding lightweight retrieval to apply such heavy
deliberation indiscriminately incurs a massive, inflexible computational
overhead. This approach ignores the "cognitive economy" of System 1,
where the majority of routine queries can be efficiently resolved
via rapid semantic recall, obviating the need for deep episodic reconstruction.

\section{Methodology}

\begin{figure*}[h] 
  \centering 
  \includegraphics[width=1\textwidth]{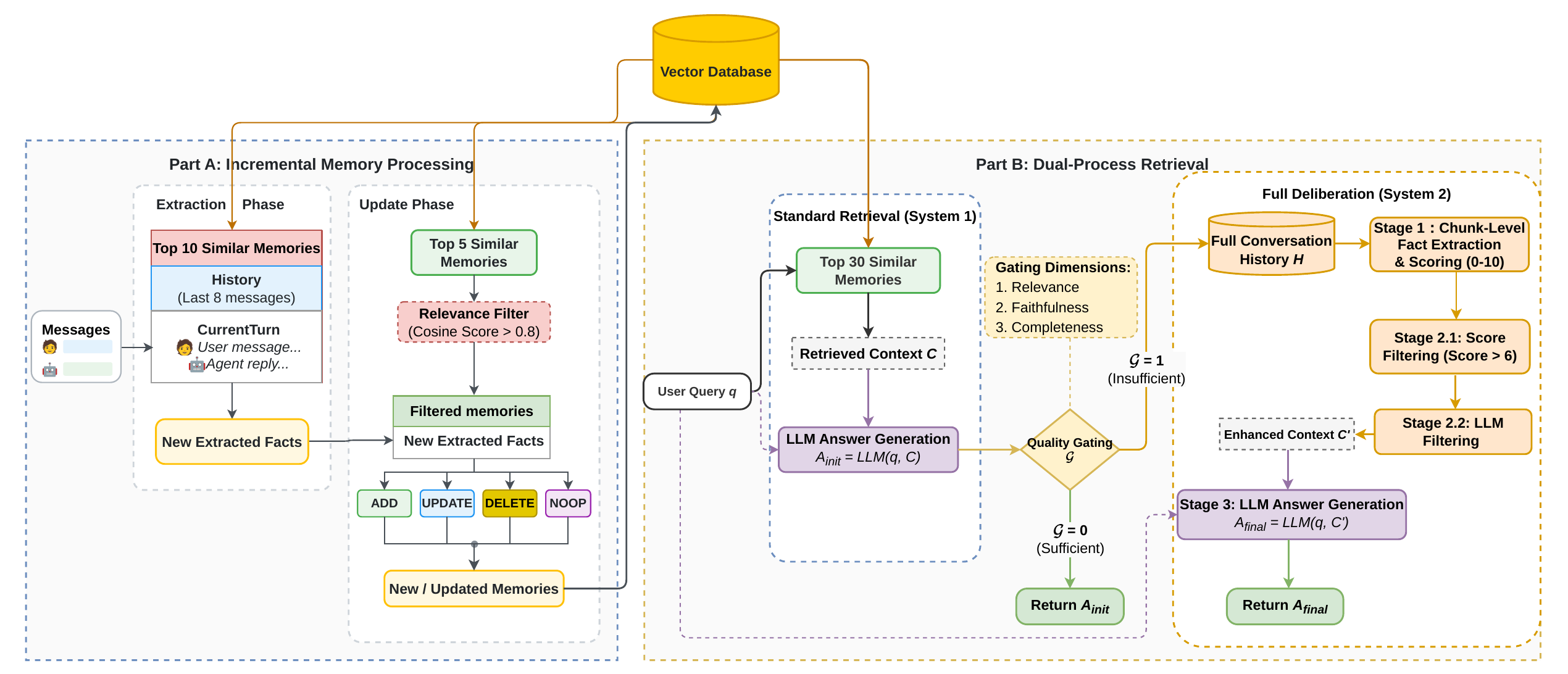} 
  \caption{Architectural overview of D-Mem. Components of 
  Part~A illustrates the memory mechanism of Mem0$^\ast$. 
  The elements highlighted in soft pale pink 
  denote our 
  key modifications over the original Mem0 
  framework: (1) utilizing the top-10 similar 
  memories rather than a general summary during 
  the Extraction Phase, and (2) incorporating an 
  additional relevance filtering step prior to 
  memory updates. Part~B illustrates the Quality 
Gating mechanism, which evaluates the initial answer
from Mem0$^\ast$ and dynamically determines whether to trigger the
Full Deliberation fallback.}
\label{fig:memory_arch} 
\end{figure*}

To circumvent the pitfalls of lossy abstraction
inherent in prior designs while maintaining
computational efficiency, we introduce D-Mem,
a dual-process memory architecture. The framework
is operationalized through three coupled components:
(1) Mem0${}^\ast$ serves as the foundational
lightweight retrieval module. It improves upon
the standard Mem0 paradigm by efficiently
extracting and updating salient conversational
memories within a vector database, designed to
rapidly resolve the majority of routine queries.
(2) The Quality Gating acts as a
dynamic evaluative router. It evaluates the initially retrieved context from 
Mem0${}^\ast$ against a multi-dimensional quality 
rubric.
Specifically, the gate assesses the retrieved
memory along three orthogonal dimensions:
\textit{Relevance}, \textit{Faithfulness \& Consistency}, and \textit{Completeness}.
If the retrieved context falls short on any of these axes,
the gate triggers the fallback mechanism.
(3) The Full
Deliberation mechanism provides a high-fidelity
fallback. Triggered exclusively when the quality
gate deems the lightweight retrieval inadequate,
this module bypasses compressed representations
to exhaustively process the raw, uncompressed
historical context. By structurally decoupling
routine semantic recall from
resource-intensive exhaustive full deliberation,
D-Mem enables high-fidelity reasoning for 
complex queries while preserving strict 
cognitive economy.

\subsection{Mem0$^\ast$: The System 1 Retrieval Foundation}

The Mem0$^\ast$ module follows an incremental processing paradigm
and serves as our foundational System 1, enabling fast associative retrieval.
As illustrated in Figure~\ref{fig:memory_arch}~Part~A,
this module comprises two phases: \textit{extraction phase} and \textit{update phase}.

Formally, to reflect the dynamic nature of
incremental processing, we model the
conversation as a continuous stream. Let
$\mathcal{H}_{t-1} = (m_1, m_2, \dots, m_{2t-2})$
denote the accumulated dialogue history prior to
the $t$-th interaction round. Each message $m_i$
represents an individual utterance augmented with
timestamp and speaker metadata. The maintenance
pipeline dynamically triggers upon the ingestion
of the new interaction round at step $t$, defined
as the message pair $(m_{2t-1}, m_{2t})$.

\begin{itemize}
  \item \textbf{Extraction Phase:} Mem0$^\ast$
        conditions the LLM on two complementary
        sources: (1) the top-$10$ most semantically similar
        existing memories $\mathcal{F}$ retrieved from the vector database,
        and (2) recent messages $(m_{2t-10}, \dots, m_{2t-2})$
        from $\mathcal{H}_{t-1}$. Here $\mathcal{F}$ serves as a
        dynamic background context and the recent
        message sequence offers granular temporal
        context. This dual contextual information,
        combined with the new message pair, forms
        a comprehensive input for the LLM to extract
        salient memories $\Omega=\{\omega_1,\dots, \omega_n\}$.
  \item \textbf{Update Phase:} Once the set of
        candidate memories $\Omega$ is extracted, the system
        retrieves the top-5 semantically similar historical
        memories for each $\omega_i \in \Omega$ via vector
        embeddings. To minimize cognitive load and mitigate
        noise, low-relevance items are explicitly discarded
        using a strict cosine similarity threshold ($> 0.8$).
        The retained historical memories constitute the
        contextual background $\mathcal{B}$.
        Subsequently, an LLM cross-references the newly
        extracted memories $\Omega$ against the contextual
        background $\mathcal{B}$ to evaluate potential
        contradictions and redundancies.
        This dictates precise memory operations
        for memories in $\mathcal{B}$ and $\Omega$:
        \texttt{ADD}, \texttt{UPDATE}, \texttt{DELETE}, or
        \texttt{NOOP}.
\end{itemize}

\paragraph{Robust Query Resolution.}
During inference, given a user query $q$,
Mem0$^\ast$ retrieves the top-30 most similar memories $C$
and generates an initial answer $A_{init} = \text{LLM}(q, C)$.

To further mitigate hallucination,
we also employ a rigorous filtering pipeline to
derive a highly refined context $C'$:

To filter out noise memories in $C$, an LLM
evaluates the initial context $C$ against the query $q$.
It retains only strictly necessary memories $C_{filter}$ to answer
$q$. In the edge case where aggressive filtering
eliminates all candidates (i.e., $C_{filter} =
  \emptyset$), the system conservatively falls back
to the original unfiltered context $C$ to
preserve recall.

\subsection{Gated Deliberation Policies}

To operationalize the routing between the Mem0$^\ast$
and the Full Deliberation fallback (as mentioned in Subsection~\ref{sec:upperbound}),
we formalize a \textbf{Gating Policy} $\mathcal{G}$.

Formally, given the initial answer $A_{init}$,
the user query $q$, and the retrieved context
$C$, we define the gating function $\mathcal{G}$:
$$\mathcal{G}: (A_{init}, q, C) \rightarrow \{0, 1\}.$$
Here, the system outputs $A_{init}$ if $\mathcal{G}(\dots) = 0$,
and it executes the
full deliberation if $\mathcal{G}(A_{init}, q, C) = 1$.

To comprehensively evaluate this dual-process
routing, we investigate three distinct
\textbf{gated deliberation policies}:
\paragraph{Policy 1: Majority Voting.}
The system generates three candidate answers from
the same Top-30 context (with temperature $>0$
for diversity). A judge determines whether a
\textit{majority consensus} exists---i.e., at
least two of the three answers are semantically
equivalent. If a majority exists, one answer is
selected from the majority group (e.g., by
preferring clarity and verbatim extraction);
otherwise $\mathcal{G}(A_{init}, q, C) = 1$ and
the system falls back to Full Deliberation.

\paragraph{Policy 2: Consensus.}
As in Majority Voting, three answers are generated
from Top-30 retrieval. The trigger requires
\textit{full} semantic agreement: all three
answers must be equivalent. If any answer differs,
$\mathcal{G}(A_{init}, q, C) = 1$ and the system
invokes Full Deliberation.
This policy is stricter than Majority Voting and
triggers fallback more often.

\paragraph{Policy 3: Quality Gating (ours).}
An LLM checks the quality of the initial answer $A_{init}$
against a quality rubric with three
dimensions: \textit{Relevance}, \textit{Faithfulness \& Consistency},
and \textit{Completeness}.
If the answer successfully passes all three dimensions, 
then the system returns $A_{init}$; otherwise,
$\mathcal{G}(A_{init}, q, C) = 1$
and the system falls back to Full Deliberation.

\subsection{Full Deliberation}
\label{sec:upperbound}

The Full Deliberation mechanism serves
both as a new baseline and as the
fallback path for all Gated Deliberation policies.
Instead of relying on semantic search, it
processes the complete conversation
history through three stages:

\paragraph{Stage 1: Chunked Fact Extraction.}
The conversation is partitioned into chunks of 60
messages each. For each chunk, an LLM
extracts query-relevant facts and assigns a
relevance score (0--10). A sliding context
window of 4 preceding messages maintains
continuity across chunks.

\paragraph{Stage 2: Multi-stage Filtering.}
Extracted facts are sorted by relevance score.
A preliminary threshold ($>6$) filters out
weakly relevant facts; if more than 6 facts
remain, an additional LLM-based filter
selects the most pertinent subset.

\paragraph{Stage 3: Answer Generation.}
The filtered facts replace the Top-30 memories $C$
in the answer generation prompt, producing
the final response.

This method is computationally expensive, which
motivates gated deliberation that
invokes it only when needed.

\section{Experiments}
\label{sec:experiments}

\subsection{Setup}
\label{sec:experimental_setup}

\paragraph{Datasets.}
We evaluate on two benchmark datasets: 

(1) \textbf{LoCoMo} \cite{maharana2024evaluating}.
It contains 10 long-term dialogues with an average length of 24K tokens. 
We utilize 1,540 questions spanning four core reasoning
categories: Single-hop, Multi-hop, Temporal,
and Open-domain (the adversarial question category 
is explicitly excluded from our evaluation).

(2) \textbf{RealTalk} \cite{lee2025realtalk}.
A real-world dialogue dataset containing 10 conversations, 
each averaging over 16,000 words. 
It features 728 questions evaluated across three categories: Multi-hop,
Temporal, and Open-domain.

These datasets primarily focus on English-language
dialogue and are designed to evaluate the agent's ability 
to maintain coherence over long-term dialogue history.
For comprehensive details regarding the demographic distribution and specific linguistic phenomena of these datasets, we refer readers to their original documentation.

\paragraph{Metrics.}
We report three complementary metrics: 
the F1 score for lexical overlap, 
the BLEU score for n-gram fidelity, and an 
LLM-as-a-Judge (hereafter, LLM) score for semantic 
equivalence. All scores are reported as 
percentages. 
For the LLM-as-a-Judge evaluation, 
we follow the binary accuracy protocol established 
by \citet{mem02025}. 
Specifically, this approach employs 
GPT-4o-mini to evaluate whether the generated 
response is semantically consistent with the ground truth. 
Rather than penalizing minor formatting differences, it provides a 
robust assessment that better correlates with human judgment by tolerating generative 
variations and relative temporal expressions.
See Appendix~\ref{app:prompt_llm_judge} for the complete prompt template.

\paragraph{LLMs.}
All methods are evaluated with GPT-4o-mini
as the primary backbone. 
We use OpenAI text-embedding-3-small as our embedding model.
We additionally report
results with Qwen3-235B-Instruct to assess 
generalization. 

\paragraph{Baselines.}
We compare our results against the following baselines: 
Full Context, which feeds the 
entire conversation history to the model; 
standard RAG, a standard retrieval-augmented generation 
approach that chunks
dialogues into 4096-token segments for dense 
retrieval;
LangMem~\cite{chase2022langchain}, Mem0~\cite{mem02025}, 
Zep~\cite{zep2024}, Nemori~\cite{nan2025nemori}, 
EMem-G~\cite{zhou2025ememg}. See \cite{nan2025nemori}
for more details on these baselines.

\subsection{Overall Performance}
Table~\ref{tab:summary_overall} summarizes the overall 
F1, LLM, and BLEU scores plus response time and token 
usage for all methods across the four experimental 
settings (see Tables~\ref{tab:overall_gpt_locomo}--\ref{tab:overall_qwen_real} for details).

\begin{table*}[t]
  \centering
  \caption{Summary of overall scores and efficiency 
  across four settings. The baseline performance comes from \cite{nan2025nemori}
and \cite{zhou2025ememg}.}
  \label{tab:summary_overall}
  \small
  \setlength{\tabcolsep}{4pt}
  \begin{tabular}{l ccccc ccccc}
    \toprule
                         & \multicolumn{5}{c}{\textbf{LoCoMo}} & \multicolumn{5}{c}{\textbf{RealTalk}}                                                                                                                                           \\
    \cmidrule(lr){2-6} \cmidrule(lr){7-11}
    \textbf{Method}      & \textbf{F1}                         & \textbf{LLM}                          & \textbf{BLEU} & \textbf{Time(s)} & \textbf{Tokens} & \textbf{F1}   & \textbf{LLM}  & \textbf{BLEU} & \textbf{Time(s)} & \textbf{Tokens} \\
    \midrule
    \multicolumn{11}{c}{\textit{GPT-4o-mini}}                                                                                                                                                                                                    \\
    \midrule
    Full Context  & 46.2 & 72.3 & 37.8 & -- & -- & -- & -- & -- & -- & -- \\
    RAG           & 20.8 & 30.2 & 16.4 & -- & -- & -- & -- & -- & -- & -- \\
    LangMem       & 35.8 & 51.3 & 29.4 & -- & -- & -- & -- & -- & -- & -- \\
    Mem0          & 41.5 & 61.3 & 34.2 & -- & -- & -- & -- & -- & -- & -- \\
    Zep           & 37.5 & 58.5 & 30.9 & -- & -- & -- & -- & -- & -- & -- \\
    Nemori       & 49.5 & 74.4 & 38.5 & -- & -- & -- & -- & -- & -- & -- \\
    \midrule
    Mem0$^\ast$           & 51.2                                & 72.7                                  & 41.0          & 1.28             & 2191            & 37.3          & 59.1          & 23.7          & 3.27             & 2303            \\
    Filter               & 51.6                                & 74.0                                  & 41.6          & 2.67             & 3190            & 38.4          & 60.7          & 24.8          & 4.14             & 3376            \\
    Majority Voting      & 51.3                                & 73.1                                  & 41.1          & 3.32             & 7534            & 37.5          & 60.4          & 23.7          & 5.40             & 8794            \\
    Consensus            & \textbf{53.5}                                & 76.1                                  & 43.0          & 9.55             & 15757           & 39.1          & 62.4          & 25.0          & 15.45            & 21949           \\
    Quality Gating(ours) & \textbf{53.5}                                & \textbf{76.3}                         & \textbf{43.1}          & 8.03             & 12681           & \textbf{39.4}          & \textbf{62.5}          & \textbf{25.4}          & 13.00            & 16786           \\
    \midrule
    Full Deliberation    & 55.3                      & 78.4                        & 44.2          & 23.73            & 35435           & 40.6 & 62.8 & 27.4 & 27.95            & 48772           \\
    \midrule
    \multicolumn{11}{c}{\textit{Qwen3-235B-Instruct}}                                                                                                                                                                                                     \\
    \midrule
    Mem0$^\ast$           & 48.1                                & 74.7                                  & 39.7          & 1.52             & 2443            & 35.4          & 62.5          & 21.3          & 2.00             & 2777            \\
    Filter               & 50.3                                & 76.8                                  & 42.0          & 2.71             & 3554            & 35.2          & 60.0          & 21.2          & 3.22             & 4119            \\
    Majority Voting      & 49.2                                & 75.4                                  & 40.8          & 4.41             & 9149            & 35.8          & 60.6          & 21.4          & 5.45             & 11009           \\
    Consensus            & \textbf{51.1}                                & 76.4                                  & 42.4          & 7.84             & 15407           & \textbf{36.3}          & 62.4          & 21.8          & 12.35            & 23673           \\
    Quality Gating(ours) & 51.0                                & \textbf{78.6}                         & \textbf{42.6}          & 8.56             & 15574           & 35.5          & \textbf{63.1}          & \textbf{22.0}          & 16.51            & 26417           \\   
    \midrule
    Full Deliberation    & 53.7                       & 78.6                                  & 45.1   & 17.03            & 39101           & 37.5 & 64.4 & 24.8 & 27.79            & 57956           \\
    \bottomrule
  \end{tabular}
\end{table*}

\paragraph{Comparison with Baselines.}
Our enhanced System 1 baseline 
(Mem0$^\ast$) achieves an F1 score of 51.2
on the LoCoMo dataset utilizing GPT-4o-mini,
significantly outperforming both the original Mem0 (41.5)
and outperforming the recently proposed Nemori (49.5).

Furthermore, by establishing an exhaustive deliberate 
reading process, the Full Deliberation mechanism
boosts the F1 score to 55.3
and the LLM score to 78.4 on the LoCoMo dataset
with GPT-4o-mini. This approach effectively mitigates
the "Lost-in-the-Middle" phenomenon,
demonstrating substantial gains over the Full Context (F1: 46.2, LLM: 72.3).
It also significantly outperforms Nemori (LLM 74.4).

In summary, while our System 1 baseline (Mem0$^\ast$) already 
achieves competitive performance, the Full 
Deliberation establishes a new upper bound.

\paragraph{Effectiveness of Dual-Process Routing.}
The core objective of D-Mem is to approximate the Full Deliberation without incurring its massive computational overhead. 
The \textit{Majority Voting} policy achieves a modest F1 
improvement over Mem0$^\ast$ (51.3 vs. 51.2) but at the 
cost of an approximate 3$\times$ token overhead (7,534 vs. 2,191 tokens). 
This suggests that Majority Voting
is prone to premature 
convergence on incorrect semantic snippets. 
Conversely, the 
\textit{Consensus} policy proves overly conservative
compared to Quality Gating: 
while it successfully elevates the F1 score to 53.5, 
it forces excessive fallbacks, resulting in heavy 
computational costs (15,757 tokens).

\paragraph{Generalization and Robustness.}
The efficacy of the Quality Gating mechanism transcends 
specific datasets and model architectures. 

On the Real-world Dialogue (RealTalk) dataset with GPT-4o-mini, the 
structural advantages of D-Mem remain substantial: 
Quality Gating achieves the better F1 score (39.4) and 
LLM-as-a-judge score (62.5) than Consensus (39.1 and 62.4). 
Mirroring the LoCoMo results, 
it successfully recovers 97.0\% of the Full Deliberation's 
F1 score (40.6) while consuming merely 34.4\% of its 
token cost (16,786 vs. 48,772). 

With the Qwen3-235B-Instruct backbone, we observe a notable 
divergence between evaluation metrics: while Quality 
Gating outperforms Consensus in LLM-as-a-judge scores 
(78.6 vs. 76.4 on LoCoMo; 63.1 vs. 62.4 on RealTalk), 
it yields slightly lower F1 scores (e.g., 51.0 vs. 51.1 on LoCoMo). 
This discrepancy is primarily attributed to the 
distinct alignment styles of the underlying model. 
Qualitative inspection reveals that Qwen3-235B-Instruct is 
inclined toward explanatory generation, 
explicitly outputting a ``step-by-step'' prior
to its final answer in 12 separate instances with Quality Gating. 
In contrast, the Consensus policy—which derives the 
final response via extraction from multiple 
candidate drafts—effectively prunes these 
deliberative steps.

\subsection{In-depth Analysis of Quality Gating}

\begin{figure*}[h] 
  \centering 
  \includegraphics[width=0.8\textwidth]{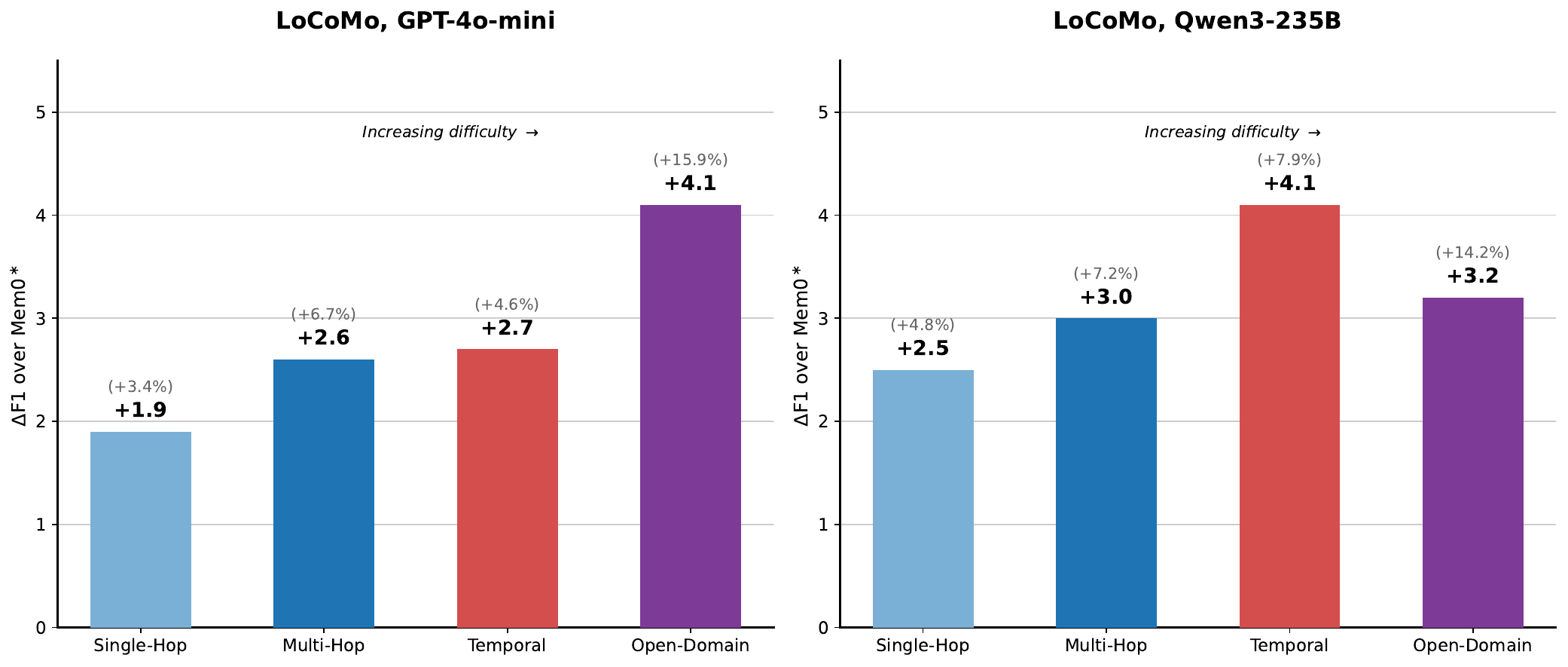} 
  \caption{F1 Improvement of Quality Gating over Mem0$^\ast$ by Question Category on the LoCoMo Dataset.}
\label{fig:f1_improvement} 
\end{figure*}

\paragraph{Per-Category Analysis}
To further understand where the Quality Gating 
mechanism yields its gains, we break down the 
F1 improvement of Quality Gating over 
Mem0$^\ast$ by question category on the LoCoMo 
dataset (Figure~\ref{fig:f1_improvement}). A clear positive correlation emerges: 
the improvement scales with question 
difficulty. For Single-Hop questions—simple 
fact lookups where the top-$K$ retrieval 
typically provides sufficient evidence—Quality 
Gating produces the smallest absolute gain 
(+1.9 F1 with GPT-4o-mini, +2.5 with 
Qwen3-235B-Instruct). For Multi-Hop questions, which 
require chaining facts scattered across 
multiple memories, the gain increases notably 
(+2.6 and +3.0, respectively), as the quality 
rubric's Completeness dimension effectively 
detects when retrieved context covers only 
a subset of the required reasoning chain. 
Temporal and Open-Domain categories benefit 
the most (+2.7/+4.1 with GPT-4o-mini; +4.1/+3.2 
with Qwen3-235B-Instruct), precisely because these 
categories suffer most severely from lossy 
abstraction: temporal queries lose relative 
time anchors during memory compression, while 
open-domain queries demand broader contextual 
grounding that top-$K$ retrieval rarely captures. 
This pronounced upward trend between question 
difficulty and Quality Gating improvement 
confirms that the gating mechanism selectively 
escalates queries where static retrieval is 
fundamentally insufficient, rather than 
applying uniform overhead across all question 
types.

\paragraph{Impact of Filtering and Model-Specific Variations.}
We observe that the explicit filtering step in 
Mem0$^\ast$ yields a consistent improvement for 
GPT-4o-mini across datasets, successfully 
mitigating the ``Lost-in-the-Middle'' phenomenon. 
However, this benefit does not generalize uniformly. 
When deploying Qwen3-235B-Instruct on the highly 
noisy RealTalk dataset, the addition of explicit 
filtering counterintuitively degrades performance compared to 
the unfiltered Mem0$^\ast$ (e.g., the LLM score drops 
from 62.5 to 60.0). We hypothesize that this 
performance inversion occurs because 
Qwen3-235B-Instruct struggles to recognize unstated 
temporal anchors or implicit logical bridges 
during the relevance assessment. 
In such noise-heavy contexts, this can lead to the 
erroneous pruning of critical context. 
This indicates that the efficacy of aggressive 
context filtering 
is dependent on the underlying model's alignment style 
and inherent discriminative capabilities.

Crucially, our proposed \textit{Quality Gating} policy 
achieves a superior balance between performance and efficiency. On the LoCoMo 
dataset with GPT-4o-mini, it achieves a strict 
improvement over the Consensus policy: it matches Consensus F1 score (53.5) and 
yields a higher LLM-as-a-judge score (76.3), 
while simultaneously reducing token consumption by 
nearly 19.5\% (12,681 vs. 15,757 tokens). Compared to the 
exhaustive fallback, Quality Gating effectively 
recovers nearly 96.7\% of the Full Deliberation's F1 
performance. Remarkably, it accomplishes this while 
utilizing only 35.8\% of the Full Deliberation's tokens 
(12,681 vs. 35,435) and approximately one-third of the 
inference latency.

\subsection{Fallback Mechanism Analysis}
\label{sec:fallback_analysis}

\begin{table*}[t]
  \centering
  \caption{Impact of Fallback Mechanism in Gated Deliberation across Datasets. W/O FB: Without Fallback; W/ FB: With Fallback; Rate: Percentage of queries falling into this routing status.}
  \label{tab:proactive_usage_merged}
  \scriptsize
  \setlength{\tabcolsep}{4pt} 
  \begin{tabular}{ll ccccc ccccc}
    \toprule
    \multirow{2}{*}{\textbf{Method}} & \multirow{2}{*}{\textbf{Status}} & \multicolumn{5}{c}{\textbf{LoCoMo}} & \multicolumn{5}{c}{\textbf{RealTalk}} \\
    \cmidrule(lr){3-7} \cmidrule(lr){8-12}
    & & \textbf{Rate} & \textbf{F1} & \textbf{LLM} & \textbf{BLEU} & \textbf{Tokens} & \textbf{Rate} & \textbf{F1} & \textbf{LLM} & \textbf{BLEU} & \textbf{Tokens} \\
    \midrule
    \multicolumn{12}{c}{\textbf{GPT-4o-mini}} \\
    \midrule
    \multirow{2}{*}{Majority Voting} 
    & W/O FB & 98.2\% & 51.6 & 73.4 & 41.4 & 6902.1  & 97.1\% & 38.0 & 61.1 & 23.9 & 7262.1  \\
    & W/ FB  & 1.8\%  & 35.1 & 59.3 & 28.7 & 41082.7 & 2.9\%  & 19.7 & 38.1 & 17.3 & 58445.0 \\
    \midrule
    \multirow{2}{*}{Consensus}       
    & W/O FB & 75.5\% & 57.4 & 78.1 & 46.7 & 6834.7  & 70.5\% & 42.1 & 65.7 & 25.9 & 7148.4  \\
    & W/ FB  & 24.5\% & 41.6 & 69.8 & 31.5 & 42315.1 & 29.5\% & 32.0 & 54.4 & 22.9 & 55959.9 \\
    \midrule
    \multirow{2}{*}{Quality Gating}  
    & W/O FB & 75.9\% & 53.8 & 78.1 & 43.4 & 4254.6  & 75.3\% & 41.6 & 66.8 & 26.4 & 4512.6  \\
    & W/ FB  & 24.1\% & 52.7 & 70.6 & 42.1 & 38582.9 & 24.7\% & 32.8 & 49.4 & 22.4 & 53218.3 \\
    \midrule
    \multicolumn{12}{c}{\textbf{Qwen3-235B-Instruct}} \\
    \midrule
    \multirow{2}{*}{Majority Voting} 
    & W/O FB & 96.4\% & 49.8 & 76.2 & 41.3 & 7610.5  & 96.0\% & 36.2 & 61.2 & 21.4 & 8609.1  \\
    & W/ FB  & 3.6\%  & 33.3 & 52.7 & 26.4 & 48192.8 & 4.0\%  & 26.6 & 44.8 & 21.1 & 65075.4 \\
    \midrule
    \multirow{2}{*}{Consensus}       
    & W/O FB & 80.8\% & 53.4 & 77.6 & 44.6 & 7548.9  & 74.2\% & 38.3 & 63.9 & 21.6 & 8557.2  \\
    & W/ FB  & 19.2\% & 41.5 & 71.6 & 33.1 & 47288.6 & 25.8\% & 30.6 & 58.0 & 22.5 & 65111.9 \\
    \midrule
    \multirow{2}{*}{Quality Gating}  
    & W/O FB & 72.1\% & 52.7 & 81.8 & 44.6 & 4728.4  & 65.0\% & 39.1 & 67.0 & 22.9 & 5431.0  \\
    & W/ FB  & 27.9\% & 46.6 & 71.6 & 37.1 & 42751.3 & 35.0\% & 28.9 & 55.7 & 20.3 & 63345.1 \\
    \bottomrule
  \end{tabular}
\end{table*}

Table~\ref{tab:proactive_usage_merged} breaks down 
the performance and resource consumption based 
on whether the fallback mechanism was triggered. 

\paragraph{Adaptability to Dataset Complexity}
Consistently across all settings, 
the fallback rate consistently scales up from LoCoMo dataset to RealTalk dataset. 
For example, on the LoCoMo dataset with Quality Gating, Qwen3-235B-Instruct triggers fallback 27.9\% of the time; 
on the substantially more challenging RealTalk dataset, 
this rate rises to 35.0\%. This dynamic behavior 
demonstrates that Quality Gating successfully 
calibrates to the varying complexity of the 
dialogue history.

\paragraph{Selection Bias in Fallback Routing.}
Consistently across all settings, queries requiring 
the fallback mechanism yield lower scores than those 
resolved in the fast path (e.g., F1 drops from 41.6 to 
32.8 on RealTalk using GPT-4o-mini). This reflects a 
natural selection bias: the quality gate effectively 
isolates the most inherently difficult queries,
thus demonstrating the effectiveness of the gating mechanism in identifying queries that are likely to fail under static retrieval.

\paragraph{Model-Specific Sensitivity in Quality Gating.}
Notably, Qwen3-235B-Instruct exhibits significantly higher fallback rates under the Quality Gating policy compared to GPT-4o-mini (e.g., 35.0\% vs. 24.7\% on RealTalk).
We attribute this to Qwen3-235B-Instruct's advanced reasoning 
capabilities, which facilitate a more rigorous assessment 
against the quality rubric. As a stricter adjudicator, 
the model more readily identifies subtle logical gaps 
or evidentiary deficiencies, ensuring the fallback 
mechanism is proactively engaged whenever static 
retrieval is insufficient.

\subsection{Case Study: Why Static Retrieval Fails}
\label{sec:error_analysis}

To better understand the underlying failure modes 
of static semantic retrieval—and why our 
D-Mem framework is necessary—we examine a 
representative error from the Temporal reasoning 
category.

\begin{quote}
  \small
  \textbf{Query:} When did Caroline go to the LGBTQ support group?\\
  \textbf{Ground Truth:} 7 May 2023\\
  \textbf{Raw Context ($\mathcal{H}$):} \textit{1:56 pm on 8 May, 2023, Caroline: I went to a LGBTQ support group yesterday and it was so powerful.}\\
  \textbf{Baseline Output:} 8 May 2023 (Retrieved Memory: \textit{``1:56 pm on 8 May, 2023: Caroline found the transgender stories at the LGBTQ support group inspiring''})\\
  \textbf{Full Deliberation Output:} 7 May 2023
\end{quote}

This failure illustrates how standard
incremental memory acts as a lossy
abstraction. Because background updates
occur without a guiding query, the LLM
over-compresses the dialogue. It retains
the thematic essence and the message's
absolute timestamp (8 May) but
permanently discards the relative
temporal arithmetic needed to resolve
``yesterday'', losing crucial
context-dependent dependencies before
retrieval even begins.

\section{Conclusion}
In this work, we first introduced Full Deliberation, 
a mechanism designed to mitigate the lossy 
abstraction of standard retrieval-based memory. 
It also mitigates the "Lost in the Middle" phenomenon 
inherent in full-context processing 
and achieves strong performance. 
To address the prohibitive computational costs of 
Full Deliberation, we further proposed D-Mem, 
a dual-process memory architecture. Through Quality 
Gating, D-Mem dynamically routes queries between 
rapid associative recall and Full Deliberation. 
Empirical evaluations demonstrate that D-Mem achieves 
near-Full Deliberation performance on the LoCoMo and 
RealTalk benchmarks, while successfully reducing both 
token consumption and inference latency.

\newpage
\section{Limitations}
\label{sec:future_work}
While Full Deliberation is a powerful mechanism for mitigating the limitations of retrieval-based memory, it inherently consumes a massive amount of input tokens by exhaustively processing the entire dialogue history. Although our Quality Gating mechanism effectively reduces these token costs, this exhaustive context scanning still limits the architecture's scalability for lifelong deployments with infinite context horizons.

Furthermore, because the current Full Deliberation mechanism splits the conversation history into isolated chunks for parallel fact extraction and relies solely on the LLM's native self-attention over the extracted isolated facts, the system lacks explicit logical chaining. This poses a significant bottleneck for long-horizon reasoning that requires global context and cross-chunk dependencies. Consequently, a state-tracking architecture that explicitly synthesizes causal and temporal logic across chunks is needed.
\bibliography{09022026}

@article{xi2023rise,
  title={The rise and potential of large language model based agents: A survey},
  author={Xi, Zhiheng and Chen, Wenxiang and Guo, Xin and He, Wei and Ding, Yiwen and Hong, Boyang and Zhang, Ming and Wang, Junzhe and Jin, Senjie and Zhou, Enyu and others},
  journal={Science China Information Sciences},
  volume={68},
  number={2},
  pages={121101},
  year={2025},
  publisher={Springer}
}

@inproceedings{
zhang2025gmemory,
title={G-Memory: Tracing Hierarchical Memory for Multi-Agent Systems},
author={Guibin Zhang and Muxin Fu and Kun Wang and Guancheng Wan and Miao Yu and Shuicheng YAN},
booktitle={The Thirty-ninth Annual Conference on Neural Information Processing Systems},
year={2025},
url={https://openreview.net/forum?id=mmIAp3cVS0}
}

@misc{wang2026emem,
      title={E-mem: Multi-agent based Episodic Context Reconstruction for LLM Agent Memory}, 
      author={Kaixiang Wang and Yidan Lin and Jiong Lou and Zhaojiacheng Zhou and Bunyod Suvonov and Jie Li},
      year={2026},
      eprint={2601.21714},
      archivePrefix={arXiv},
      primaryClass={cs.AI},
      url={https://arxiv.org/abs/2601.21714}, 
}

@misc{yan2025gam,
      title={General Agentic Memory Via Deep Research}, 
      author={B. Y. Yan and Chaofan Li and Hongjin Qian and Shuqi Lu and Zheng Liu},
      year={2025},
      eprint={2511.18423},
      archivePrefix={arXiv},
      primaryClass={cs.CL},
      url={https://arxiv.org/abs/2511.18423}, 
}

@article{wang2024survey,
  title={A survey on large language model based autonomous agents},
  author={Wang, Lei and Ma, Chen and Feng, Xueyang and Zhang, Zeyu and Yang, Hao and Zhang, Jingsen and Chen, Zhiyuan and Tang, Jiakai and Chen, Xu and Lin, Yankai and others},
  journal={Frontiers of Computer Science},
  volume={18},
  number={6},
  pages={186345},
  year={2024},
  publisher={Springer}
}

@article{liu2024lost,
    title = "Lost in the Middle: How Language Models Use Long Contexts",
    author = "Liu, Nelson F.  and
      Lin, Kevin  and
      Hewitt, John  and
      Paranjape, Ashwin  and
      Bevilacqua, Michele  and
      Petroni, Fabio  and
      Liang, Percy",
    journal = "Transactions of the Association for Computational Linguistics",
    volume = "12",
    year = "2024",
    address = "Cambridge, MA",
    publisher = "MIT Press",
    url = "https://aclanthology.org/2024.tacl-1.9/",
    doi = "10.1162/tacl_a_00638",
    pages = "157--173"
}

@misc{packer2023memgpt,
      title={MemGPT: Towards LLMs as Operating Systems}, 
      author={Charles Packer and Sarah Wooders and Kevin Lin and Vivian Fang and Shishir G. Patil and Ion Stoica and Joseph E. Gonzalez},
      year={2024},
      eprint={2310.08560},
      archivePrefix={arXiv},
      primaryClass={cs.AI},
      url={https://arxiv.org/abs/2310.08560}, 
}

@misc{nan2025nemori,
      title={Nemori: Self-Organizing Agent Memory Inspired by Cognitive Science}, 
      author={Jiayan Nan and Wenquan Ma and Wenlong Wu and Yize Chen},
      year={2025},
      eprint={2508.03341},
      archivePrefix={arXiv},
      primaryClass={cs.AI},
      url={https://arxiv.org/abs/2508.03341}, 
}

@misc{zhou2025ememg,
      title={A Simple Yet Strong Baseline for Long-Term Conversational Memory of LLM Agents}, 
      author={Sizhe Zhou and Jiawei Han},
      year={2025},
      eprint={2511.17208},
      archivePrefix={arXiv},
      primaryClass={cs.CL},
      url={https://arxiv.org/abs/2511.17208}, 
}

@misc{mem02025,
      title={Mem0: Building Production-Ready AI Agents with Scalable Long-Term Memory}, 
      author={Prateek Chhikara and Dev Khant and Saket Aryan and Taranjeet Singh and Deshraj Yadav},
      year={2025},
      eprint={2504.19413},
      archivePrefix={arXiv},
      primaryClass={cs.CL},
      url={https://arxiv.org/abs/2504.19413}, 
}

@article{evans2013dual,
author = {Jonathan St. B. T. Evans and Keith E. Stanovich},
title ={Dual-Process Theories of Higher Cognition: Advancing the Debate},

journal = {Perspectives on Psychological Science},
volume = {8},
number = {3},
pages = {223-241},
year = {2013},
doi = {10.1177/1745691612460685},
    note ={PMID: 26172965},

URL = { 
    
        https://doi.org/10.1177/1745691612460685
    
    

},
eprint = { 
    
        https://doi.org/10.1177/1745691612460685
}
}

@inproceedings{maharana2024evaluating,
    title = "Evaluating Very Long-Term Conversational Memory of {LLM} Agents",
    author = "Maharana, Adyasha  and
      Lee, Dong-Ho  and
      Tulyakov, Sergey  and
      Bansal, Mohit  and
      Barbieri, Francesco  and
      Fang, Yuwei",
    editor = "Ku, Lun-Wei  and
      Martins, Andre  and
      Srikumar, Vivek",
    booktitle = "Proceedings of the 62nd Annual Meeting of the Association for Computational Linguistics (Volume 1: Long Papers)",
    month = aug,
    year = "2024",
    address = "Bangkok, Thailand",
    publisher = "Association for Computational Linguistics",
    url = "https://aclanthology.org/2024.acl-long.747/",
    doi = "10.18653/v1/2024.acl-long.747",
    pages = "13851--13870"
}

@misc{lee2025realtalk,
      title={REALTALK: A 21-Day Real-World Dataset for Long-Term Conversation}, 
      author={Dong-Ho Lee and Adyasha Maharana and Jay Pujara and Xiang Ren and Francesco Barbieri},
      year={2025},
      eprint={2502.13270},
      archivePrefix={arXiv},
      primaryClass={cs.CL},
      url={https://arxiv.org/abs/2502.13270}, 
}

@misc{chase2022langchain,
  author       = {Harrison Chase},
  title        = {{LangChain}},
  year         = {2022},
  publisher    = {GitHub},
  journal      = {GitHub repository},
  howpublished = {\url{https://github.com/langchain-ai/langchain}},
  note         = {Accessed: 2025-07-20}
}

@inproceedings{
asai2024,
title={Self-{RAG}: Learning to Retrieve, Generate, and Critique through Self-Reflection},
author={Akari Asai and Zeqiu Wu and Yizhong Wang and Avirup Sil and Hannaneh Hajishirzi},
booktitle={The Twelfth International Conference on Learning Representations},
year={2024},
url={https://openreview.net/forum?id=hSyW5go0v8}
}

@book{kahneman2011thinking,
  title={Thinking, fast and slow},
  author={Kahneman, Daniel},
  year={2011},
  publisher={Farrar, Straus and Giroux},
  address={New York}
}

@misc{zhu2026,
      title={From Lossy to Verified: A Provenance-Aware Tiered Memory for Agents}, 
      author={Qiming Zhu and Shunian Chen and Rui Yu and Zhehao Wu and Benyou Wang},
      year={2026},
      eprint={2602.17913},
      archivePrefix={arXiv},
      primaryClass={cs.DB},
      url={https://arxiv.org/abs/2602.17913}, 
}

@inproceedings{
amem,
title={A-Mem: Agentic Memory for {LLM} Agents},
author={Wujiang Xu and Zujie Liang and Kai Mei and Hang Gao and Juntao Tan and Yongfeng Zhang},
booktitle={The Thirty-ninth Annual Conference on Neural Information Processing Systems},
year={2025},
url={https://openreview.net/forum?id=FiM0M8gcct}
}

@misc{zep2024,
      title={Zep: A Temporal Knowledge Graph Architecture for Agent Memory}, 
      author={Preston Rasmussen and Pavlo Paliychuk and Travis Beauvais and Jack Ryan and Daniel Chalef},
      year={2025},
      eprint={2501.13956},
      archivePrefix={arXiv},
      primaryClass={cs.CL},
      url={https://arxiv.org/abs/2501.13956}, 
}

@inproceedings{zhong2023memorybank,
author = {Zhong, Wanjun and Guo, Lianghong and Gao, Qiqi and Ye, He and Wang, Yanlin},
title = {MemoryBank: enhancing large language models with long-term memory},
year = {2024},
isbn = {978-1-57735-887-9},
publisher = {AAAI Press},
url = {https://doi.org/10.1609/aaai.v38i17.29946},
doi = {10.1609/aaai.v38i17.29946},
booktitle = {Proceedings of the Thirty-Eighth AAAI Conference on Artificial Intelligence and Thirty-Sixth Conference on Innovative Applications of Artificial Intelligence and Fourteenth Symposium on Educational Advances in Artificial Intelligence},
articleno = {2198},
numpages = {8},
series = {AAAI'24/IAAI'24/EAAI'24}
}

@inproceedings{activerag,
  title={{ThinkNote}: Enhancing Knowledge Integration and Utilization of Large Language Models via Constructivist Cognition Modeling},
  author={Xu, Zhipeng and Liu, Zhenghao and Yan, Yukun and Wang, Shuo and Yu, Shi and Zeng, Zheni and Xiao, Chaojun and Liu, Zhiyuan and Yu, Ge and Xiong, Chenyan},
  booktitle={Findings of the Association for Computational Linguistics: EACL 2026},
  year={2026},
  note={To appear}
}

@inproceedings{query_expansion,
author = {Zhang, Wenjing and Liu, Zhaoxiang and Wang, Kai and Lian, Shiguo},
title = {Query Expansion and Verification with Large Language Model for Information Retrieval},
year = {2024},
isbn = {978-981-97-5671-1},
publisher = {Springer-Verlag},
address = {Berlin, Heidelberg},
url = {https://doi.org/10.1007/978-981-97-5672-8_29},
doi = {10.1007/978-981-97-5672-8_29},
booktitle = {Advanced Intelligent Computing Technology and Applications: 20th International Conference, ICIC 2024, Tianjin, China, August 5–8, 2024, Proceedings, Part IV},
pages = {341–351},
numpages = {11},
location = {Tianjin, China}
}

@inproceedings{
reranking,
title={Rank{RAG}: Unifying Context Ranking with Retrieval-Augmented Generation in {LLM}s},
author={Yue Yu and Wei Ping and Zihan Liu and Boxin Wang and Jiaxuan You and Chao Zhang and Mohammad Shoeybi and Bryan Catanzaro},
booktitle={The Thirty-eighth Annual Conference on Neural Information Processing Systems},
year={2024},
url={https://openreview.net/forum?id=S1fc92uemC}
}

@inproceedings{filtering,
author = {Rossi, Nicholas and Lin, Juexin and Liu, Feng and Yang, Zhen and Lee, Tony and Magnani, Alessandro and Liao, Ciya},
title = {Relevance Filtering for Embedding-based Retrieval},
year = {2024},
isbn = {9798400704369},
publisher = {Association for Computing Machinery},
address = {New York, NY, USA},
url = {https://doi.org/10.1145/3627673.3680095},
doi = {10.1145/3627673.3680095},
booktitle = {Proceedings of the 33rd ACM International Conference on Information and Knowledge Management},
pages = {4828–4835},
numpages = {8},
location = {Boise, ID, USA},
series = {CIKM '24}
}

@inproceedings{
reprompting,
title={Take a Step Back: Evoking Reasoning via Abstraction in Large Language Models},
author={Huaixiu Steven Zheng and Swaroop Mishra and Xinyun Chen and Heng-Tze Cheng and Ed H. Chi and Quoc V Le and Denny Zhou},
booktitle={The Twelfth International Conference on Learning Representations},
year={2024},
url={https://openreview.net/forum?id=3bq3jsvcQ1}
}

@inproceedings{consensus,
author = {Du, Yilun and Li, Shuang and Torralba, Antonio and Tenenbaum, Joshua B. and Mordatch, Igor},
title = {Improving factuality and reasoning in language models through multiagent debate},
year = {2024},
publisher = {JMLR.org},
booktitle = {Proceedings of the 41st International Conference on Machine Learning},
articleno = {467},
numpages = {31},
location = {Vienna, Austria},
series = {ICML'24}
}

\newpage
\begin{appendices}
  \section{Prompt Templates}
  \label{app:prompts}

In this section, we provide the complete prompt templates used in our experiments. 
For the sake of reproducibility and fair comparison, 
the prompts utilized for the static retrieval baseline and the evaluation judge are sourced from the \texttt{mem0} open-source repository \cite{mem02025}. Templates specifically designed for our Multi-dimensional Quality Gating and Full Deliberation modules are denoted with ``(Ours)''.

  \subsection{Answer Generation Prompt (Mem0)}
  \label{app:prompt_answer}

  The following prompt is used for generating answers from retrieved memories in all methods:

  \begin{lstlisting}
You are an intelligent memory assistant tasked with retrieving 
accurate information from conversation memories.

# CONTEXT:
You have access to memories from two speakers in a conversation. 
These memories contain timestamped information that may be 
relevant to answering the question.

# INSTRUCTIONS:
1. Carefully analyze all provided memories from both speakers
2. Pay special attention to the timestamps to determine the answer
3. If the question asks about a specific event or fact, look for 
   direct evidence in the memories
4. If the memories contain contradictory information, prioritize 
   the most recent memory
5. If there is a question about time references (like "last year", 
   "two months ago", etc.), calculate the actual date based on the 
   memory timestamp. For example, if a memory from 4 May 2022 
   mentions "went to India last year," then the trip occurred in 2021.
6. Always convert relative time references to specific dates, months, 
   or years. For example, convert "last year" to "2022" or "two 
   months ago" to "March 2023" based on the memory timestamp. 
   Ignore the reference while answering the question.
7. Focus only on the content of the memories from both speakers. 
   Do not confuse character names mentioned in memories with the 
   actual users who created those memories.
8. The answer should be less than 5-6 words.
9. If the memory information is insufficient to answer the question, 
   respond with "NONE".

# APPROACH (Think step by step):
1. First, examine all memories that contain information related 
   to the question
2. Examine the timestamps and content of these memories carefully
3. Look for explicit mentions of dates, times, locations, or events 
   that answer the question
4. If the answer requires calculation (e.g., converting relative 
   time references), show your work
5. Formulate a precise, concise answer based solely on the evidence 
   in the memories
6. Double-check that your answer directly addresses the question asked
7. Ensure your final answer is specific and avoids vague time references

Memories for user {{speaker_1_user_id}}:
{{speaker_1_memories}}

Memories for user {{speaker_2_user_id}}:
{{speaker_2_memories}}

Question: {{question}}

Answer:
\end{lstlisting}

  \subsection{Filter Prompt (Ours)}
  \label{app:prompt_filter}

  This prompt is used in the Filter method to filter irrelevant memories:

  \begin{lstlisting}
You are an AI assistant that selects which memory items are 
relevant to the user's question.

Instructions:
- Only output the indexes of memory items that are relevant 
  to the question.
- If none are relevant, output an empty list [].
- Do not answer the question.
- Be strict: only select items that clearly help answer the question.

User question: "{user_question}"

Memory items:
1. {memory_1}
2. {memory_2}
3. {memory_3}
...

Output format (JSON array of indexes):
[ ... ]
\end{lstlisting}

  \subsection{Fact Extraction Prompt (Ours)}
  \label{app:prompt_extraction}

  This prompt is used in the Full Deliberation method to extract relevant facts from conversation history:

  \begin{lstlisting}
You are a Personal Information Organizer, specialized in accurately 
extracting and scoring facts from conversations.

# ROLE:
- Your job is to extract **relevant facts** from a given conversation.
- Each extracted fact must directly support answering the user's question.

# INPUT:
- A user question
- **History**: Previous conversation messages for context. 
  *DO NOT* extract from this section.
- **Conversation**: Current conversation chunk to extract facts from. 
  **Your task is to extract new facts learned *only* from this section** 
  based on user question and History.

# INSTRUCTIONS:
1. Extract only the facts from **Conversation** that can **help answer 
   the given question**. Ignore irrelevant parts.
2. The extracted fact **must be a complete, standalone statement**. 
   - Include **timestamp and speaker** for every fact.
   - **Resolve pronouns** (like "it", "they") using the conversation context.
   - Pay attention to **contextual meaning**.
3. Assign a **relevance score (0-10)**:
   - 0 = Irrelevant
   - 5-6 = Somewhat relevant
   - 7-8 = Clearly relevant
   - 9-10 = Essential for answering the question
4. Output must be in **strict JSON format** as a *list* of *objects*, 
   wrapped under a single "facts" key.
   Each object in the list must contain **EXACTLY TWO KEYS**:
   - "fact": (string) The resolved statement with timestamp/speaker.
   - "score": (integer) The relevance score (0-10).
   - Only include facts with a **score >= 5**.
   - If no fact qualifies, return {"facts": []}.
\end{lstlisting}

  \subsection{Fact Filtering System Prompt (Ours)}
  \label{app:prompt_filter_facts}

  This prompt is used in the Full Deliberation method to further filter extracted facts:

  \begin{lstlisting}
You are an AI Relevance Filter. Your task is to analyze a list of 
numbered memory items and select *only* the indexes of the items 
that are strictly necessary and relevant to answer a user's question.

# Instructions:
- Return a JSON object with a single key "indexes", which contains 
  an array of the relevant memory indexes.
- **IMPORTANT: Indexes start from 1. The first item is index 1, 
  the second is index 2, and so on.**
- Select memory items that:
  1. Provide facts, dates, events, or context that directly help 
     answer the question.
  2. Contain key details for a correct and complete answer.
  3. Be strict: only memory items that clearly help answer the 
     question should be selected.
- Avoid:
  - Memory items unrelated to the question.
  - Vague or duplicate entries (if two items are almost the same, 
    keep only one).
- If no relevant memory exists, return {"indexes": []}.
- **Do not answer the question.**

# Output Format
Return *only* a valid JSON object with an "indexes" key.
Example: {"indexes": [1, 3]}
\end{lstlisting}

  \subsection{Consensus Prompt (Ours)}
  \label{app:prompt_full_agree}

  This prompt is used in the Consensus policy to check if all three generated answers are semantically equivalent:

  \begin{lstlisting}
You are a Semantic Consistency Judge.
Task: Analyze a Question and 3 Candidate Answers to determine if 
all answers are semantically equivalent.

# Judgement Criteria:
1. **True Match (Equivalent):**
   - **Formatting:** "2023-01-01" == "Jan 1, 2023".
   - **Abbreviation:** "NYC" == "New York City".
   - **Structure:** "$50" == "It is $50".
   - **Context:** Use the 'Question' to resolve ambiguities.

2. **List/Set Logic (Multiple Items):**
   - **Order Independence:** "Red, Blue" == "Blue, Red" (Unless 
     the question explicitly asks for a ranked order).
   - **Delimiters:** "Apple, Banana" == "Apple and Banana".
   - **Completeness:** All items must be present. "A, B" != "A, B, C" 
     (Mismatch).

3. **Mismatch (Not Equivalent):**
   - **Contradiction:** "Yes" != "No".
   - **Different Values:** "100m" != "1km".
   - **Partial/Missing:** "Paris" != "Paris, Texas" (if specific 
     location is needed).
   - **Subset/Superset:** If one answer has 3 items and another 
     has 2, it is a mismatch.

4. **Selection Logic (If Match):**
   - **STRICT EXTRACTION:** You MUST choose one answer **VERBATIM** 
     from the provided 'Answers' list.
   - **Priority Hierarchy (Apply in order):**
   - **Step A (Strip Fillers):** Avoid answers containing 
     conversational fillers like "It is", "The answer is", 
     "I think", "Confirmed".
   - **Step B (Maximize Clarity):** Between the remaining options, 
     prefer the one with **standard formatting** over extreme 
     abbreviations or ambiguity.
   - **Step C (Tie-Breaker):** If both are clear and have no fillers, 
     select the **Shortest** one.
   - **NO EDITING:** Do not remove words.

# Output Format (MUST be valid JSON):
- If ALL 3 match:
  {"consensus": true, "answer": "answer"}
- If ANY differ:
  {"consensus": false}

# Constraints:
- Output ONLY valid JSON. NO Markdown, NO extra text.
- When consensus is true, "answer" MUST be one of the provided 
  Answers strings verbatim.
\end{lstlisting}

  \subsection{Majority Voting Prompt (Ours)}
  \label{app:prompt_maj_vote}
This prompt is used in the Majority Voting policy 
to check if at least 2 out of 3 generated answers are 
semantically equivalent:
  \begin{lstlisting}
You are a Majority Consistency Judge.
Task: Analyze a Question and 3 Answers. Determine if there is a 
**majority consensus** (at least 2 out of 3 answers share the same 
semantic meaning).

# Judgement Logic:
1. **Semantic Grouping:**
   - Compare the meanings of Answer 1, 2, and 3.
   - "Jan 2023" == "January 2023" (Match).
   - "Paris" != "London" (Mismatch).

2. **Majority Rule:**
   - **3/3 Match:** All three are equivalent. -> Consensus FOUND.
   - **2/3 Match:** Two answers are equivalent, one is an outlier. 
     -> Consensus FOUND (Ignore the outlier).
   - **0 Matches:** All three are different. -> Consensus NOT FOUND.

# Selection Logic (If Consensus Found):
1. **Filter:** Consider ONLY the answers that form the majority group. 
   (Discard the outlier).
2. **Strict Extraction:** You MUST select one answer **VERBATIM** 
   (word-for-word) from the majority group.
3. **Priority Hierarchy (Apply in order):**
   - **Step A (Strip Fillers):** Avoid answers containing 
     conversational fillers like "It is", "The answer is", 
     "I think", "Confirmed".
   - **Step B (Maximize Clarity):** Between the remaining options, 
     prefer the one with **standard formatting** over extreme 
     abbreviations or ambiguity.
   - **Step C (Tie-Breaker):** If both are clear and have no fillers, 
     select the **Shortest** one.

# Output Format (MUST be valid JSON):
- If Majority exists:
  {"consensus": true, "answer": "<one original answer string 
   verbatim from Answers>"}
- If NO Majority:
  {"consensus": false}

# Constraints:
- Output ONLY valid JSON. NO Markdown, NO extra text.
- When consensus is true, "answer" MUST be one of the provided 
  Answers strings verbatim.
- **NEVER** pick the outlier answer.
\end{lstlisting}

  \subsection{Quality Gating Prompt (Ours)}
  \label{app:prompt_rubric}

  This prompt is used in the Quality Gating policy 
  to evaluate answer quality and trigger 
  Full Deliberation:

  \begin{lstlisting}
You are the **Answer Quality Auditor**. Your task is to act as a strict gatekeeper for a RAG system.

### Input Data
# memories
{{speaker_1_memories}}
{{speaker_2_memories}}
# question
{{question}}
# answer
{{answer}}

### Evaluation Protocol (Execute in Order)

#### 1. Relevance
- **Check**: Does the Answer directly and fully address the Question?
- **Rule**:
  - **Irrelevant**: Off-topic? -> FAIL.
  - **Vagueness**: If memories are vague and Answer admits it -> PASS.
  - **Retrieval Trigger**: If the answer is information is missing, mark as FAIL.
- **Output**: {"relevance": "pass" | "fail"}

#### 2. Faithfulness & Consistency (Fact Check)
- **Check**: Does the Answer contradict the Memories?
- **Rule**:
  - **Hallucination**: Does it contain info NOT in memories? -> FAIL.
  - **Conflict**: Does it contradict the latest memory timestamp? -> FAIL.
  - **Logic Separation**: Do NOT mark math/time errors here (handle them in Logic Check).
- **Output**: {"faithfulness": "pass" | "fail"}

#### 3. Completeness & Relevance
- **Check**: Does the Answer directly and fully address the Question?
- **Rule**:
  - **Irrelevant**: Off-topic? -> FAIL.
  - **Incomplete**: Misses key details requested? -> FAIL.
  - **Vagueness**: If memories are vague and Answer admits it -> PASS.
  - **Retrieval Trigger**: If the answer is "I don't know" or information is missing, mark as FAIL.
- **Output**: {"completeness_relevance": "pass" | "fail"}

### Output Format (JSON)
Return ONLY valid JSON.
{
    "relevance": {
        "status": "pass" | "fail"
    },
    "faithfulness": {
        "status": "pass" | "fail"
    },
    "completeness_relevance": {
        "status": "pass" | "fail"
    }
}
\end{lstlisting}

\subsection{LLM-as-a-Judge Prompt (Mem0)}
\label{app:prompt_llm_judge}
This prompt  is used to evaluate the quality of generated answers using an LLM as a judge:

\begin{lstlisting}
Your task is to label an answer to a question as 'CORRECT' or 'WRONG'. You will be given the following data:
    (1) a question (posed by one user to another user), 
    (2) a 'gold' (ground truth) answer, 
    (3) a generated answer
which you will score as CORRECT/WRONG.

The point of the question is to ask about something one user should know about the other user based on their prior conversations.
The gold answer will usually be a concise and short answer that includes the referenced topic, for example:
Question: Do you remember what I got the last time I went to Hawaii?
Gold answer: A shell necklace
The generated answer might be much longer, but you should be generous with your grading - as long as it touches on the same topic as the gold answer, it should be counted as CORRECT. 

For time related questions, the gold answer will be a specific date, month, year, etc. The generated answer might be much longer or use relative time references (like "last Tuesday" or "next month"), but you should be generous with your grading - as long as it refers to the same date or time period as the gold answer, it should be counted as CORRECT. Even if the format differs (e.g., "May 7th" vs "7 May"), consider it CORRECT if it's the same date.

Now it's time for the real question:
Question: {question}
Gold answer: {gold_answer}
Generated answer: {generated_answer}

First, provide a short (one sentence) explanation of your reasoning, then finish with CORRECT or WRONG. 
Do NOT include both CORRECT and WRONG in your response, or it will break the evaluation script.

Just return the label CORRECT or WRONG in a json format with the key as "label".
\end{lstlisting}

\section{Implementation Details}
  \label{app:implementation}

  \subsection{Hyperparameter Settings}
  \label{app:hyperparameters}

  Key hyperparameters used in our experiments:

  \begin{itemize}
    \item \textbf{TopK}: 30 (number of memories retrieved per speaker)
    \item \textbf{Temperature for Answer Generation}: 0.0 (deterministic)
    \item \textbf{Temperature for Multiple Answers}: 0.7 (for diversity in Majority Voting and Consensus methods)
    \item \textbf{MESSAGES\_CHUNK\_SIZE}: 60 (messages per chunk in Full Deliberation method)
    \item \textbf{HISTORY\_SIZE}: 4 (previous messages as context in Full Deliberation method)
    \item \textbf{Preliminary Score Threshold}: 6 (minimum relevance score for filtering in Full Deliberation method)
    \item \textbf{LLM Filter Threshold}: 6 (minimum facts to trigger LLM-based filtering)
    \item \textbf{Fact Extraction Score Threshold}: 5 (minimum relevance score for extraction in Full Deliberation method)
  \end{itemize}

  \subsection{System Configuration}
  \label{app:system_config}

  \begin{itemize}
    \item \textbf{Vector Store}: Qdrant (local instance)
    \item \textbf{Embedding Model}: OpenAI's \texttt{text-embedding-3-small}
    \item \textbf{API Providers}: OpenAI (GPT-4o-mini), Qwen (Qwen3-235B-Instruct)
  \end{itemize}

  \section{Comprehensive Evaluation Results}
\label{sec:detailed_results}

In the main text, due to space constraints, we 
present the aggregated performance and key 
analytical visualizations of our proposed D-Mem 
framework. In this section, we provide the 
complete, fine-grained evaluation results across all tested dimensions, datasets, and base models. 

\begin{itemize}
    \item \textbf{Table \ref{tab:overall_gpt_locomo}}: Full results on the LoCoMo dataset using GPT-4o-mini. This table also includes results from prior baseline frameworks (LangMem, Mem0, RAG, Zep, Nemori) for a comprehensive historical comparison.
    \item \textbf{Table \ref{tab:overall_gpt_real}}: Full results on the RealTalk dataset using GPT-4o-mini. Note that RealTalk inherently excludes Single-Hop questions.
    \item \textbf{Table \ref{tab:overall_qwen_locomo}}: Full results on the LoCoMo dataset using the Qwen3-235B-Instruct model.
    \item \textbf{Table \ref{tab:overall_qwen_real}}: Full results on the RealTalk dataset using the Qwen3-235B-Instruct model.
\end{itemize}

These detailed tables substantiate the core claims made in Section \ref{sec:experiments} (Main Paper), particularly demonstrating the consistent superiority of our Multi-dimensional Quality Gating policy in balancing high-fidelity reasoning with cognitive economy.

  \begin{table*}[t]
    \centering
    \caption{Overall Performance Comparison (LoCoMo, GPT-4o-mini)}
    \label{tab:overall_gpt_locomo}
    \setlength{\tabcolsep}{3.5pt}
    \resizebox{\textwidth}{!}{
      \begin{tabular}{l ccccc ccccc ccccc ccc}
        \toprule
        \multirow{2}{*}{\textbf{Method}}            &
        \multicolumn{5}{c}{\textbf{F1 Score}}       &
        \multicolumn{5}{c}{\textbf{LLM-as-a-Judge}} &
        \multicolumn{5}{c}{\textbf{BLEU}}           &
        \multicolumn{3}{c}{\textbf{Efficiency}}                                                                                                                                                                                                                                                                                \\
        \cmidrule(lr){2-6} \cmidrule(lr){7-11} \cmidrule(lr){12-16} \cmidrule(lr){17-19}
                                                    & \textbf{S-H}  & \textbf{M-H}  & \textbf{Temp} & \textbf{O-D}  & \textbf{Avg}
                                                    & \textbf{S-H}  & \textbf{M-H}  & \textbf{Temp} & \textbf{O-D}  & \textbf{Avg}
                                                    & \textbf{S-H}  & \textbf{M-H}  & \textbf{Temp} & \textbf{O-D}  & \textbf{Avg}
                                                    & \textbf{In}   & \textbf{Out}  & \textbf{Time}                                                                                                                                                                                                                            \\
        \midrule
        Full Context                                & 53.1          & 35.4          & 44.1          & 24.5          & 46.2          & 83.0          & 66.8          & 56.2          & 48.6          & 72.3          & 44.7          & 26.1          & 36.1          & 17.2          & 37.8          & --      & --    & --     \\
        LangMem                                     & 38.8          & 33.5          & 31.9          & 29.4          & 35.8          & 61.4          & 52.4          & 24.9          & 47.6          & 51.3          & 33.1          & 23.9          & 26.2          & 23.5          & 29.4          & --      & --    & --     \\
        Mem0                                        & 44.4          & 34.3          & 44.4          & 27.1          & 41.5          & 68.1          & 60.3          & 50.4          & 40.6          & 61.3          & 37.7          & 25.2          & 37.6          & 19.4          & 34.2          & --      & --    & --     \\
        RAG                                         & 22.2          & 18.6          & 19.5          & 19.0          & 20.8          & 32.0          & 31.3          & 23.7          & 32.6          & 30.2          & 18.6          & 11.7          & 15.7          & 13.5          & 16.4          & --      & --    & --     \\
        Zep                                         & 39.7          & 27.5          & 44.8          & 22.9          & 37.5          & 63.2          & 50.5          & 58.9          & 39.6          & 58.5          & 33.7          & 19.3          & 38.1          & 15.7          & 30.9          & --      & --    & --     \\
        Nemori                               & 54.4          & 36.5          & 56.7          & 20.8          & 49.5          & 82.1          & 65.3          & 71.0          & 44.8          & 74.4          & 43.2          & 25.6          & 46.6          & 15.1          & 38.5          & --      & --    & --     \\
        \midrule
        \multicolumn{19}{l}{\textit{Basic Methods}}                                                                                                                                                                                                                                                                            \\
        Mem0$^\ast$                                      & 55.2          & 38.8          & 59.3          & 25.8          & 51.2          & 79.2          & 63.8          & 73.2          & 40.6          & 72.7          & 45.5          & 27.0          & 48.3          & 19.4          & 41.1          & 2186.2  & 5.2   & 1.278  \\
        Filter                             & 54.3          & 41.0          & 61.0          & 27.9          & 51.6          & 78.4          & 67.4          & 75.7          & 50.0          & 74.0          & 45.2          & 29.4          & 48.8          & 20.8          & 41.6          & 3161.7  & 28.5  & 2.673  \\
        \midrule
        \multicolumn{19}{l}{\textit{Gated Deliberation}}                                                                                                                                                                                                                                                                       \\
        Majority Voting                             & 54.5          & 40.2          & 60.2          & 25.1          & 51.3          & 79.1          & 65.6          & 72.0          & 46.9          & 73.1          & 45.1          & 27.9          & 49.2          & 18.4          & 41.1          & 7501.4  & 32.6  & 3.318  \\
        Consensus                                   & 57.5          & 42.2          & 61.5          & 25.3          & 53.5          & 83.5          & 68.4          & 72.9          & 44.8          & 76.1          & 47.3          & 30.3          & 49.6          & 19.9          & 43.0          & 15543.6 & 213.0 & 9.549  \\
        Quality Gating (ours) & 57.1 & 41.4 & 62.0 & 29.9 & 53.5  & 83.0 & 68.8 & 73.2 & 50.0 & 76.3 & 47.2 & 29.8 & 50.3 & 22.4 & 43.1 & 12524.6 & 156.6 & 8.03 \\
        \midrule
        Full Deliberation                           & 58.9          & 44.2 & 63.0 & 31.1 & 55.3 & 83.1          & 73.4 & 77.9 & 54.2 & 78.4 & 48.2          & 32.0 & 50.6 & 24.0 & 44.2          & 34805.0 & 629.9 & 23.725 \\
        \bottomrule
      \end{tabular}
    }
  \end{table*}

  \begin{table*}[t]
    \centering
    \caption{Overall Performance Comparison (RealTalk, GPT-4o-mini)}
    \label{tab:overall_gpt_real}
    \setlength{\tabcolsep}{3.5pt}
    \resizebox{\textwidth}{!}{
      \begin{tabular}{l cccc cccc cccc ccc}
        \toprule
        \multirow{2}{*}{\textbf{Method}}            &
        \multicolumn{4}{c}{\textbf{F1 Score}}       &
        \multicolumn{4}{c}{\textbf{LLM-as-a-Judge}} &
        \multicolumn{4}{c}{\textbf{BLEU}}           &
        \multicolumn{3}{c}{\textbf{Efficiency}}                                                                                                                                                                                                              \\
        \cmidrule(lr){2-5} \cmidrule(lr){6-9} \cmidrule(lr){10-13} \cmidrule(lr){14-16}
                                                    & \textbf{M-H}  & \textbf{Temp} & \textbf{O-D}  & \textbf{Avg}
                                                    & \textbf{M-H}  & \textbf{Temp} & \textbf{O-D}  & \textbf{Avg}
                                                    & \textbf{M-H}  & \textbf{Temp} & \textbf{O-D}  & \textbf{Avg}
                                                    & \textbf{In}   & \textbf{Out}  & \textbf{Time}                                                                                                                                                          \\
        \midrule
        \multicolumn{16}{l}{\textit{Basic Methods}}                                                                                                                                                                                                          \\
        Mem0$^\ast$                                       & 32.0          & 47.5          & 21.7          & 37.3          & 51.2          & 68.0          & 54.6          & 59.1 & 23.5          & 26.1          & 17.0 & 23.7          & 2296.5  & 6.0   & 3.272  \\
        Filter                             & 33.2          & 48.3          & 24.0          & 38.4          & 54.8          & 68.7          & 53.7          & 60.7 & 24.6          & 27.3          & 17.8 & 24.8          & 3345.1  & 31.3  & 4.140  \\
        \midrule
        \multicolumn{16}{l}{\textit{Gated Deliberation}}                                                                                                                                                                                                     \\
        Majority Voting                             & 31.6          & 48.2          & 22.2          & 37.5          & 53.2          & 69.6          & 53.7          & 60.4 & 22.6          & 27.0          & 17.3 & 23.7          & 8738.5  & 55.2  & 5.404  \\
        Consensus                                   & 34.4          & 49.1          & 22.8          & 39.1          & 57.5          & 69.6          & 54.6          & 62.4 & 24.4          & 28.2          & 17.3 & 25.0          & 21563.9 & 385.2 & 15.446 \\
        Quality Gating (ours)                             & 33.6          & 50.6          & 22.7          & 39.4          & 55.2          & 70.9 & 58.3 & 62.5          & 24.3          & 29.0          & 17.7          & 25.4          & 16555.2 & 231.1 & 13.003 \\
        \midrule
        Full Deliberation                           & 36.4          & 50.1          & 24.4 & 40.6 & 59.5 & 68.0          & 56.5          & 62.8 & 27.9 & 30.0 & 18.3 & 27.4 & 47955.0 & 816.5 & 27.951 \\
        \bottomrule
      \end{tabular}
    }
  \end{table*}

  \begin{table*}[t]
    \centering
    \caption{Overall Performance Comparison (LoCoMo, Qwen3-235B-Instruct)}
    \label{tab:overall_qwen_locomo}
    \setlength{\tabcolsep}{3.5pt}
    \resizebox{\textwidth}{!}{
      \begin{tabular}{l ccccc ccccc ccccc ccc}
        \toprule
        \multirow{2}{*}{\textbf{Method}}            &
        \multicolumn{5}{c}{\textbf{F1 Score}}       &
        \multicolumn{5}{c}{\textbf{LLM-as-a-Judge}} &
        \multicolumn{5}{c}{\textbf{BLEU}}           &
        \multicolumn{3}{c}{\textbf{Efficiency}}                                                                                                                                                                                                                                                                                \\
        \cmidrule(lr){2-6} \cmidrule(lr){7-11} \cmidrule(lr){12-16} \cmidrule(lr){17-19}
                                                    & \textbf{S-H}  & \textbf{M-H}  & \textbf{Temp} & \textbf{O-D}  & \textbf{Avg}
                                                    & \textbf{S-H}  & \textbf{M-H}  & \textbf{Temp} & \textbf{O-D}  & \textbf{Avg}
                                                    & \textbf{S-H}  & \textbf{M-H}  & \textbf{Temp} & \textbf{O-D}  & \textbf{Avg}
                                                    & \textbf{In}   & \textbf{Out}  & \textbf{Time}                                                                                                                                                                                                                            \\
        \midrule
        Mem0 (Baseline)                             & 38.0          & 29.5          & 42.5          & 16.3          & 36.0          & 58.0          & 52.5          & 49.8          & 37.5          & 54.0          & 31.7          & 20.1          & 35.1          & 13.1          & 29.1          & 1977.9  & 5.7   & 0.702  \\
        \midrule
        \multicolumn{19}{l}{\textit{Basic Methods}}                                                                                                                                                                                                                                                                            \\
        Mem0$^\ast$                                       & 51.7          & 41.5          & 51.8          & 22.6          & 48.1          & 79.9          & 74.8          & 65.7          & 59.4          & 74.7          & 45.4          & 32.1          & 38.1          & 18.0          & 39.7          & 2431.2  & 11.6  & 1.518  \\
        Filter                             & 52.8          & 41.6          & 59.1 & 24.5          & 50.3          & 79.8          & 74.5          & 77.0 & 57.3          & 76.8          & 46.7          & 32.5          & 44.6          & 20.3          & 42.0          & 3521.8  & 32.5  & 2.713  \\
        \midrule
        \multicolumn{19}{l}{\textit{Gated Deliberation}}                                                                                                                                                                                                                                                                       \\
        Majority Voting                             & 52.6          & 41.3          & 54.9          & 23.8          & 49.2          & 80.5          & 75.9          & 67.0          & 57.3          & 75.4          & 46.2          & 32.4          & 40.6          & 18.2          & 40.8          & 9059.9  & 89.0  & 4.407  \\
        Consensus                                   & 54.6          & 43.4          & 56.8          & 23.4          & 51.1          & 81.9          & 75.9          & 67.9          & 58.3          & 76.4          & 48.0          & 33.8          & 42.6          & 17.3          & 42.4          & 15187.2 & 219.3 & 7.840  \\
        Quality Gating (ours)                             & 54.2          & 44.5          & 55.9          & 25.8          & 51.0          & 83.7 & 78.7 & 70.1          & 61.5 & 78.6 & 47.8          & 34.6          & 42.4          & 21.7          & 42.6          & 15319.4 & 254.7 & 8.560  \\
        \midrule
        Full Deliberation                           & 56.6          & 47.7          & 58.8          & 28.6 & 53.7 & 83.6          & 78.7          & 71.7          & 57.3          & 78.6          & 49.6 & 38.2 & 45.3 & 25.0 & 45.1 & 38488.9 & 611.6 & 17.032 \\
        \bottomrule
      \end{tabular}
    }
  \end{table*}

  \begin{table*}[t]
    \centering
    \caption{Overall Performance Comparison (RealTalk, Qwen3-235B-Instruct)}
    \label{tab:overall_qwen_real}
    \setlength{\tabcolsep}{3.5pt}
    \resizebox{\textwidth}{!}{
      \begin{tabular}{l cccc cccc cccc ccc}
        \toprule
        \multirow{2}{*}{\textbf{Method}}            &
        \multicolumn{4}{c}{\textbf{F1 Score}}       &
        \multicolumn{4}{c}{\textbf{LLM-as-a-Judge}} &
        \multicolumn{4}{c}{\textbf{BLEU}}           &
        \multicolumn{3}{c}{\textbf{Efficiency}}                                                                                                                                                                                                                                 \\
        \cmidrule(lr){2-5} \cmidrule(lr){6-9} \cmidrule(lr){10-13} \cmidrule(lr){14-16}
                                                    & \textbf{M-H}  & \textbf{Temp} & \textbf{O-D}  & \textbf{Avg}
                                                    & \textbf{M-H}  & \textbf{Temp} & \textbf{O-D}  & \textbf{Avg}
                                                    & \textbf{M-H}  & \textbf{Temp} & \textbf{O-D}  & \textbf{Avg}
                                                    & \textbf{In}   & \textbf{Out}  & \textbf{Time}                                                                                                                                                                             \\
        \midrule
        \multicolumn{16}{l}{\textit{Basic Methods}}                                                                                                                                                                                                                             \\
        Mem0$^\ast$                                       & 32.1          & 44.1          & 18.6          & 35.4          & 59.1          & 68.0          & 55.6          & 62.5          & 26.1          & 18.8          & 15.4          & 21.3          & 2765.5  & 11.0   & 1.995  \\
        Filter                             & 32.1          & 43.3          & 19.8          & 35.2          & 56.8          & 65.8          & 51.8          & 60.0          & 24.9          & 19.7          & 15.4          & 21.2          & 4075.5  & 43.9   & 3.215  \\
        \midrule
        \multicolumn{16}{l}{\textit{Gated Deliberation}}                                                                                                                                                                                                                        \\
        Majority Voting                             & 32.8          & 44.8 & 17.9          & 35.8          & 55.8          & 67.7          & 52.8          & 60.6          & 26.5          & 18.9          & 14.8          & 21.4          & 10858.4 & 150.7  & 5.446  \\
        Consensus                                   & 33.6          & 44.4          & 20.1          & 36.3          & 59.8          & 68.0          & 52.8          & 62.4          & 26.5          & 19.2          & 16.5          & 21.8          & 23162.0 & 510.5  & 12.353 \\
        Quality Gating (ours)                      & 33.3          & 43.1          & 19.2          & 35.5          & 61.5          & 67.7          & 53.7          & 63.1          & 27.5          & 19.3          & 14.8          & 22.0          & 25716.8 & 700.2  & 16.515 \\
        \midrule
        Full Deliberation                           & 35.4 & 44.6          & 22.6 & 37.5 & 61.8 & 68.7          & 59.3 & 64.4 & 29.1 & 23.4 & 17.3 & 24.8 & 56703.4 & 1253.0 & 27.792 \\
        \bottomrule
      \end{tabular}
    }
  \end{table*}

  \section{Computational Resources and Reproducibility}
  To ensure the reproducibility of our experimental results, we provide details regarding the computational environment and the models utilized in this study:

\begin{itemize}
    \item \textbf{Infrastructure}: All experimental orchestration, data pre-processing, and local evaluation scripts were executed on a \textbf{Lenovo laptop} equipped with an \textbf{NVIDIA GeForce RTX 4070 Laptop GPU (8GB VRAM)}, and an Intel Core i9-13900HX CPU.
    
    \item \textbf{Model Access}: We utilized proprietary models via official API endpoints to ensure consistency:
    \begin{itemize}
    \item \textbf{GPT-4o-mini}: Accessed via the OpenAI API. Its exact parameter count remains proprietary and has not been disclosed by the provider.
    \item \textbf{Qwen-235B-Instruct}: Accessed via the Alibaba DashScope API. It is a Mixture-of-Experts (MoE) model with a total of 235 billion parameters and 22 billion active parameters.
\end{itemize}

    \item \textbf{Computational Budget}: The entire evaluation process for D-Mem is highly efficient; all experimental runs were completed within 24 wall-clock hours through parallel API invocations, involving a total consumption of approximately 400 million tokens.
    \item \textbf{Software and Evaluation Packages}: Our implementation and evaluation framework was built upon the following technical stack: 
\begin{itemize}
    \item \textbf{Orchestration}: Python 3.10 with the \texttt{openai} (v2.7.1) library for model inference and API orchestration. We also utilized \texttt{langchain} (v1.2.7) for auxiliary memory management tasks.
    \item \textbf{Tokenization}: \texttt{tiktoken} (v0.12.0) with the \texttt{cl100k\_base} encoding was employed for precise token counting and to ensure compliance with model-specific context window constraints.
    \item \textbf{Evaluation Metrics}: Lexical overlap metrics were computed using \texttt{nltk} (v3.9.2) for F1 and BLEU scores, and \texttt{rouge-score} (v0.1.2) for ROUGE-L. Semantic evaluations (LLM-as-a-Judge) were executed by GPT-4o-mini, following the multi-dimensional rubric detailed in Appendix~\ref{app:prompts}. Statistical analysis and visualization were performed using \texttt{scikit-learn} (v1.7.2), \texttt{matplotlib} (v3.10.7), and \texttt{pandas} (v2.3.3).
\end{itemize}

    \item \textbf{Statistical Transparency}: Due to the substantial computational costs associated with processing the extensive contexts in the LoCoMo and RealTalk benchmarks, all performance metrics reported in this paper are derived from a \textbf{single, exhaustive execution} of the evaluation pipeline. To ensure the reliability and reproducibility of these results, we utilized \textbf{deterministic decoding} (e.g., setting $\texttt{temperature} = 0$) for memory retrieval, except for the Majority Voting and Consensus.

  \end{itemize}

  \section{Ethics Statement}
\label{sec:ethics}

In accordance with the ACL Code of Ethics, we acknowledge and discuss the potential risks and broader impacts associated with the deployment of long-term memory systems for LLM agents like D-Mem.

\paragraph{Privacy and Data Security.} 
The core capability of D-Mem involves persistently storing and retrieving extensive user interaction histories. This inherently introduces risks related to data privacy in real-world applications, especially if the conversational context contains Personally Identifiable Information (PII) or sensitive operational data. To mitigate these concerns during our research phase, we strictly evaluated our framework on publicly available benchmark datasets (e.g., LoCoMo and RealTalk). We verified that these standard benchmarks have been appropriately pre-processed and anonymized by their creators to remove PII and mitigate offensive content. However, for future real-world deployment, practitioners must implement strict data encryption and allow users to actively manage or delete their memory states.
\paragraph{Environmental and Computational Impact.} 
While our Quality Gating mechanism successfully mitigates redundant compute for simple queries (System 1), the Full Deliberation module (System 2) requires exhaustive context processing. As demonstrated in our efficiency metrics, this exhaustive nature increases token consumption and inference latency. Large-scale deployment of such dual-process systems could lead to a substantial carbon footprint. Future work should explore more eco-friendly deliberation alternatives, such as deploying smaller, specialized language models for the gating functions and Full Deliberation.

\paragraph{Memory-Induced Bias and Safety.} 
A highly retentive memory system runs the risk of perpetuating or amplifying historical biases. If an agent ingests toxic or factually incorrect statements from a user, these ``poisoned'' memories could be retrieved during future multi-hop reasoning, leading to unsafe or hallucinated outputs over time. We urge developers to pair D-Mem with robust safety guardrails and memory-sanitization protocols before user-facing deployment.

\section{Artifact Licenses and Terms of Use}
\label{app:licenses}

To ensure responsible NLP research and compliance with intellectual property guidelines, we outline the licenses of the scientific artifacts used and created in this work:

\begin{itemize}
    \item \textbf{Utilized Datasets}: The LoCoMo and RealTalk datasets are used strictly for academic evaluation purposes, adhering to their respective open-source distribution terms (e.g., CC BY 4.0).
    \item \textbf{Utilized Models and Frameworks}: We accessed GPT-4o-mini via the official OpenAI API under their terms of service. The Qwen3-235B-Instruct model via the official Qwen API under their terms of service. The baseline memory framework, Mem0, is distributed under the Apache License 2.0.
    \item \textbf{Created Artifacts}: The source code for our D-Mem framework, along with all evaluation scripts, is distributed under the MIT License. The intended use of our created artifacts is to facilitate reproducibility and future academic research, which is entirely compatible with the original licenses of the utilized data and frameworks.
  \end{itemize}
\end{appendices}

\end{document}